\documentclass{article}

    \PassOptionsToPackage{numbers, compress}{natbib}

     \usepackage[preprint]{neurips_2020}

\usepackage[utf8]{inputenc} 
\usepackage[T1]{fontenc}    
\usepackage{hyperref}       
\usepackage{url}            
\usepackage{booktabs}       
\usepackage{amsfonts}       
\usepackage{nicefrac}       
\usepackage{microtype}      
\usepackage{rotating}
\usepackage{multirow}
\usepackage{longtable}
\usepackage{pdflscape}
\usepackage{array}
\usepackage{amssymb}
\usepackage{enumitem}
\usepackage[dvipsnames]{xcolor}
\newcolumntype{I}{!{\vrule width 1pt}}

\title{Towards Standard Criteria for human evaluation of Chatbots: A Survey}

\author{%
  Hongru Liang \\
  Sichuan University\\
  China \\
  \texttt{lianghongru@scu.edu.cn} \\
   \And
  Huaqing Li \\
  University of Science and Technology of China \\
  China \\
  \texttt{redcobain1003@gmail.com} \\

}

\begin{document}
\maketitle
\begin{abstract}
Human evaluation is becoming  a necessity to test the performance of Chatbots. However, off-the-shelf settings suffer the severe reliability and replication issues partly because of the extremely high diversity of criteria. It is high time to come up with standard criteria and exact definitions. To this end, we conduct a through investigation of 105 papers involving human evaluation for Chatbots. Deriving from this, we propose five standard criteria along with precise definitions. 
\end{abstract}

\section{Introduction}
Chatbots, a.k.a., non-task-oriented dialogue systems, chit-chat with human on open-domains. Due to the lacking of specific goals~(e.g., booking a restaurant ~\cite{lei2018sequicity,jin2018explicit}, making a recommendation~\cite{lei2020estimation,lei2020interactive,li2020seamlessly,lei2020conversational,gao2021advances}, answering a question~\cite{pan2019recent,li2020molweni,zhu2021retrieving}), it is not possible for non-task-oriented dialogue systems to employ the automated goal-matching evaluation metrics~(e.g, task success rate~\cite{lei2018sequicity,lei2020estimation,lei2020interactive} and entity inform score~\cite{lei2018sequicity,zhu2021retrieving}), which are favored by task-oriented dialogue systems. The remaining automated corpus-based metrics~(e.g., BLEU~\cite{papineni2002bleu}, ROUGE~\cite{lin2004rouge}, embedding similarity score~\cite{liang2020pirhdy,lei2021have}) are also not suitable, as they are weakly correlated with the human judgements~\cite{liu2016not}. Hence, it is essential to invite human trials in the evaluation of non-task-oriented dialogue systems.\par

However, off-the-shelf approaches are criticized a lot for the poor reliability and replication. These problems are caused by the high variety in settings, especially the setting of criteria that can differ evaluation approaches from the beginning.  Moreover, we observe that part of current papers misuse the criteria. For example, \citet{gao2020dialogue} claim that they use ``\texttt{fluent}'' in human evaluation, but describe it as ``\texttt{the grammatical correctness of responses}''. Another observation is that some papers just report they use human evaluation yet lack definitions of the criteria. This suggests that standard criteria must be defined in this community.\par
To this end, we keep our eyes on the criteria used in current human evaluation of chatbots and make a through survey of 105 related papers from 2016 to 2020. Considering the above-mentioned problems in current papers, we also explore the definitions of criteria in dictionaries education papers, linguistic papers, etc. Specifically, we classify all found criteria into seven groups based on the definitions. Further, we suggest five standard criteria along with the exact definitions for human evaluation in chatbots.
\section{Data Collection}
We begin our survey by searching papers published on six top-tier natural language processing conferences from 2016 to 2020. The conferences involve Annual Meeting of the Association for Computational Linguistics~(ACL), Conference on Empirical Methods in Natural Language Processing (EMNLP), North American Chapter of the Association for Computational Linguistics (NAACL), International Conference on Computational Linguistics~(COLING), Annual Meeting of the Special Interest Group on Discourse and Dialogue~(SIGDIAL), and International Natural Language Generation Conference~(INLG). We also checked papers published on other well-known conferences, e.g., Conference on Neural Information Processing Systems~\cite{ghandeharioun2019approximating}, AAAI Conference on Artificial Intelligence~\cite{tao2018ruber}, ACM CHI Conference on Human Factors in Computing Systems~\cite{santhanam2020studying}, etc. All mentioned criteria can be found in the above six conferences. Hence, it is comprehensive enough to do investigations on ACL, EMNLP, NAACL, COLING, SIGDIAL, and INLG.\par
We conduct a two-step filtering: in the first step, we keep papers whose titles conclude ``\texttt{dialog}'', ``\texttt{conversation}'', ``\texttt{chat}'', and ``\texttt{response}''; in the second step, we read the kept papers quickly and remove those not on chatbots, those without human evaluation, and those with human evaluation but without mention of criteria. In this way, we collect 105 related papers, whose statistics is reported in Table~\ref{tab:data}. 

\begin{table}[t]
    \centering
    \begin{tabular}{cccccccc}
    \hline
         & h5 index & 2016 & 2017 & 2018 & 2019 & 2020 & Total\\
         \hline
         ACL&135 &0&2&3&13&15&33\\
         \hline
         EMNLP&112 &2&5&8&9&17&41\\
         \hline
         NAACL&90 &3&0&2&7&0&12\\
         \hline
         COLING&49 &1&0&4&0&4&9\\
         \hline
         SIGDIAL&29 &0&0&0&1&4&5\\
         \hline
         INLG&18 &0&0&0&4&1&5\\
         \hline
         Total&- &6&9&17&34&41&105\\
         \hline
    \end{tabular}
    \caption{Data statistics w.r.t. conference names, h5 indexes, years and paper numbers}
    \label{tab:data}
\end{table}

\section{Taxonomy}
We also display the detailed information of the collected papers in Table~\ref{tab:survey}. It statistically proves the high diversity of criteria in this community. First, there are more than 27 criteria in total, only a few paper use the exactly same criteria. Second, different criteria are also different in terms of popularity --- 40 papers favor the ``\texttt{Overall Quality}'', whereas only one paper uses the ``\texttt{Adequacy}''. An interesting finding is that some criteria share similar definitions in some aspects. For example, \citet{gao2020dialogue} define ``\texttt{fluent}'' as ``\texttt{the grammatical correctness~(grammaticality) of responses}''; \citet{qiu2019training} define ``\texttt{readability}'' as ``\texttt{Grammaticality}'' too. As such, towards a clearer understanding of these criteria, we classify them into 7 groups based on the similarities in definitions:
\begin{enumerate}[label={\texttt Group \arabic*}]
    \item \texttt{Fluency, Grammaticality, Correctness, Readability, Understandable};
    \item \texttt{Relevance, Coherence, Consistency, Sensibleness, Listening, Maintain Context, Logic};
    \item \texttt{Informativeness, Diversity, Specificity, Proactivity, Flexible;}
    \item \texttt{Overall Quality, Appropriateness, Naturalness, Humanness, Adequacy;}
    \item \texttt{Engagement, Interestingness;}
    \item \texttt{Empathy, Emotion;}
    \item \texttt{Others}.
\end{enumerate}
Specifically, \texttt{Group 1} involves criteria referring to the writing quality of the generated text, \texttt{Group 2} involves criteria referring to the dialogue flow, \texttt{Group 3} involves criteria referring to the details of the generated text, \texttt{Group 4} involve criteria referring to the general opinion about the generated text, \texttt{Group 5} involve criteria referring to human experience over the dialogue agent, \texttt{Group 6} involve criteria referring to human emotional feelings, and \texttt{Group 7} involve criteria that cannot be included in \texttt{Group 1-6}, e.g., Inquisitiveness~\cite{see2019makes}.

\begin{landscape}
    \begin{figure} 
    \centering  
    \includegraphics[width=\linewidth]{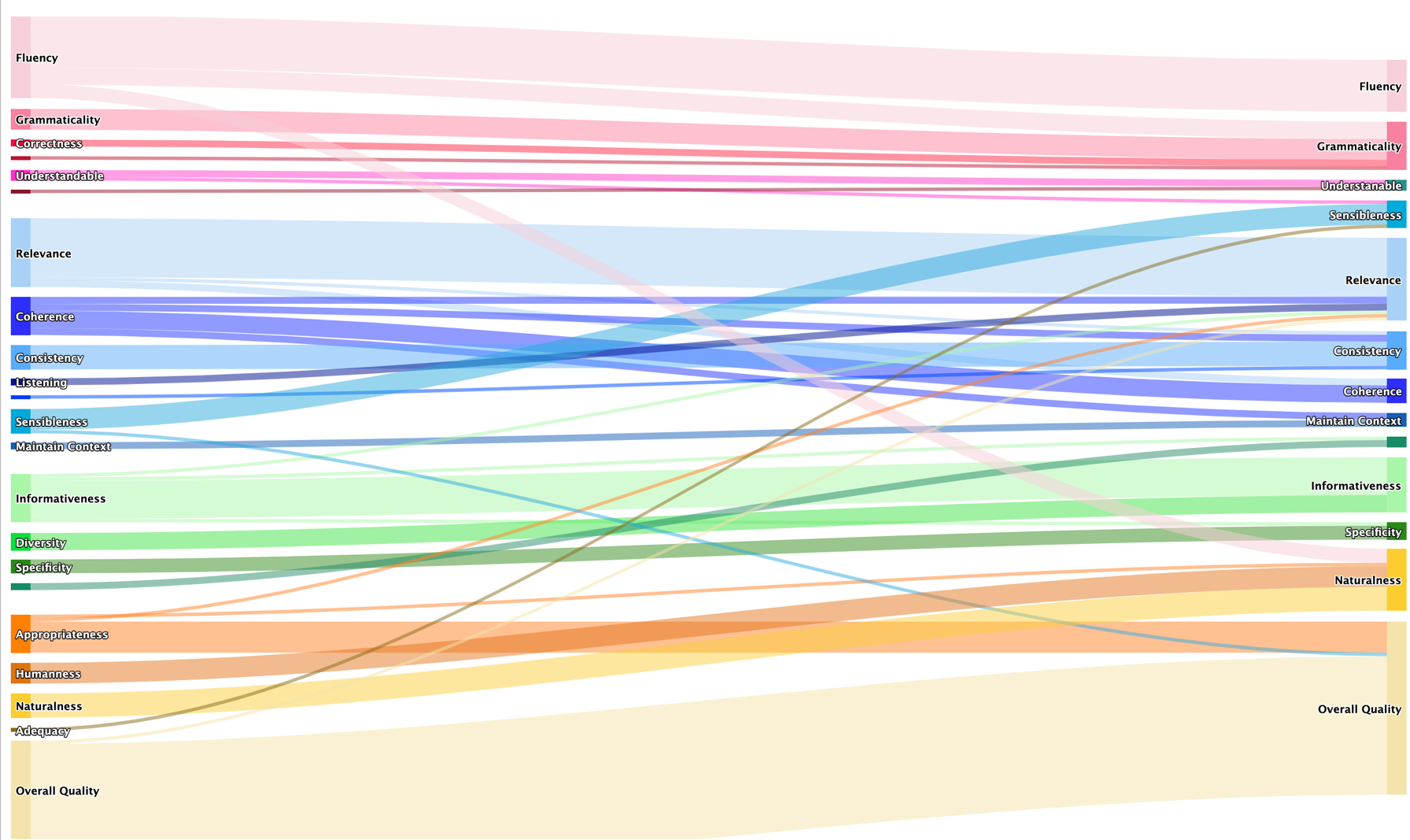} 
    \caption{Criteria associated with their actual criteria}  
    \label{Fig.sankey}  
    \end{figure}
\end{landscape}

Another finding is that many criteria are misused. For example,  \citet{gao2020dialogue} and \citet{qiu2019training} misuse ``\texttt{Fluency}'' and ``\texttt{Readability}'' as ``\texttt{Grammaticality}''. We illustrate this issue visually in Figure~\ref{Fig.sankey}. In particular, although ``\texttt{Fluency}'' ranks the second in terms of popularity over other criteria, nearly half actually are ``\texttt{Grammaticality}'' or ``\texttt{Naturalness}''. Another example is ``\texttt{Coherence}'', half of which are ``\texttt{Relevance}'', ``\texttt{Consistency}'', or  ``\texttt{Maintain Context}'' in fact. Moreover, part 
\begin{landscape}
\scriptsize
\begin{longtable}{Il|c|c|c|cIp{0.5em}|p{0.5em}|p{0.5em}|p{0.5em}|p{0.5em}Ip{0.5em}|p{0.5em}|p{0.5em}|p{0.5em}|p{0.5em}|p{0.5em}|p{0.5em}Ip{0.5em}|p{0.5em}|p{0.5em}|p{0.5em}|p{0.5em}Ip{0.5em}|p{0.5em}|p{0.5em}|p{0.5em}|p{0.5em}Ip{0.5em}|p{0.5em}Ip{0.5em}|p{0.5em}Ip{0.5em}I}

\caption{Detailed information of collected papers}
    \label{tab:survey}\\
    \hline
    \multicolumn{4}{Ic}{}  & {\textcolor{blue}{Group ID}}& \multicolumn{5}{cI}{\textcolor{blue}{1}}                        & \multicolumn{7}{cI}{\textcolor{blue}{2}  }                              & \multicolumn{5}{cI}{\textcolor{blue}{3} }               &  \multicolumn{5}{cI}{\textcolor{blue}{4} }&       \multicolumn{2}{cI}{\textcolor{blue}{5}} & \multicolumn{2}{cI}{\textcolor{blue}{6} }& \textcolor{blue}{7} \\
    \hline
     \multicolumn{4}{Ic}{} & {\begin{sideways}\textcolor{blue}{ \texttt Criteria}\end{sideways}} & {\begin{sideways}\textcolor{blue}{ \texttt Fluency}\end{sideways}} & {\begin{sideways}\textcolor{blue}{ \texttt Grammaticality}\end{sideways}} & {\begin{sideways}\textcolor{blue}{ \texttt Correctness}\end{sideways}} & {\begin{sideways}\textcolor{blue}{ \texttt Readability}\end{sideways}} & {\begin{sideways}\textcolor{blue}{ \texttt Understandable}\end{sideways}} & {\begin{sideways}\textcolor{blue}{ \texttt Relevance}\end{sideways}} & {\begin{sideways}\textcolor{blue}{ \texttt Coherence}\end{sideways}} & {\begin{sideways}\textcolor{blue}{ \texttt Consistency}\end{sideways}} & {\begin{sideways}\textcolor{blue}{ \texttt Sensibleness}\end{sideways}} & {\begin{sideways}\textcolor{blue}{ \texttt Listening}\end{sideways}} & {\begin{sideways}\textcolor{blue}{ \texttt Maintain Context}\end{sideways}} & {\begin{sideways}\textcolor{blue}{ \texttt Logic}\end{sideways}} & \begin{sideways}\textcolor{blue}{ \texttt Informativeness}\end{sideways} & {\begin{sideways}\textcolor{blue}{ \texttt Diversity}\end{sideways}} & {\begin{sideways}\textcolor{blue}{ \texttt Specificity}\end{sideways}} & {\begin{sideways}\textcolor{blue}{ \texttt Proactivity}\end{sideways}} & {\begin{sideways}\textcolor{blue}{ \texttt Flexible}\end{sideways}} & {\begin{sideways}\textcolor{blue}{ \texttt Overall Quality}\end{sideways}} & {\begin{sideways}\textcolor{blue}{ \texttt Appropriateness}\end{sideways}} & \begin{sideways}\textcolor{blue}{ \texttt Naturalness}\end{sideways} & {\begin{sideways}\textcolor{blue}{ \texttt Humanness}\end{sideways}} & {\begin{sideways}\textcolor{blue}{ \texttt Adequacy}\end{sideways}} & {\begin{sideways}\textcolor{blue}{ \texttt Engagingness}\end{sideways}} & {\begin{sideways}\textcolor{blue}{ \texttt Interestingness}\end{sideways}} & {\begin{sideways}\textcolor{blue}{ \texttt Empathy}\end{sideways}} & {\begin{sideways}\textcolor{blue}{ \texttt Emotion}\end{sideways}} & {\begin{sideways}\textcolor{blue}{ \texttt Others}\end{sideways}} \\
    \hline
    Paper & {Venue} & \multicolumn{1}{m{3.5em}|}{Citation Number} & {year} & \multicolumn{1}{m{3.5em}I}{Criteria Number} & \multicolumn{1}{p{0.5em}|}{\color{blue} 34} & \multicolumn{1}{p{0.5em}|}{\color{blue} 8} & \multicolumn{1}{p{0.5em}|}{\color{blue} 3} & \multicolumn{1}{p{0.5em}|}{\color{blue} 2} & \multicolumn{1}{p{0.5em}I}{3} &  \multicolumn{1}{p{0.5em}|}{\color{blue} 29} & \multicolumn{1}{p{0.5em}|}{\color{blue} 18} & \multicolumn{1}{p{0.5em}|}{\color{blue} 11} & \multicolumn{1}{p{0.5em}|}{\color{blue} 6} & \multicolumn{1}{p{0.5em}|}{\color{blue} 2} & \multicolumn{1}{p{0.5em}|}{\color{blue} 1} & \multicolumn{1}{p{0.5em}I}{1} & \multicolumn{1}{p{0.5em}|}{\color{blue} 26} & \multicolumn{1}{p{0.5em}|}{\color{blue} 14} & \multicolumn{1}{p{0.5em}|}{\color{blue} 3} & \multicolumn{1}{p{0.5em}|}{\color{blue} 2} & \multicolumn{1}{p{0.5em}I}{1} & \multicolumn{1}{p{0.5em}|}{\color{blue} 40} & \multicolumn{1}{p{0.5em}|}{\color{blue} 12} & \multicolumn{1}{p{0.5em}|}{\color{blue} 7} & \multicolumn{1}{p{0.5em}|}{\color{blue} 5} & \multicolumn{1}{p{0.5em}I}{1} & \multicolumn{1}{p{0.5em}|}{\color{blue} 11} & \multicolumn{1}{p{0.5em}I}{9} & \multicolumn{1}{p{0.5em}|}{\color{blue} 6} & \multicolumn{1}{p{0.5em}I}{3} & \multicolumn{1}{p{0.5em}I}{4} \\
    \hline
    \endfirsthead
    
    \multicolumn{32}{l}%
{{\bfseries \tablename\ \thetable{} -- continued from previous page}} \\
\hline
    \multicolumn{4}{Ic}{}  & {\textcolor{blue}{Group ID}}& \multicolumn{5}{cI}{\textcolor{blue}{1}}                        & \multicolumn{7}{cI}{\textcolor{blue}{2}  }                              & \multicolumn{5}{cI}{\textcolor{blue}{3} }               &  \multicolumn{5}{cI}{\textcolor{blue}{4} }&       \multicolumn{2}{cI}{\textcolor{blue}{5}} & \multicolumn{2}{cI}{\textcolor{blue}{6} }& \textcolor{blue}{7} \\
    \hline
     \multicolumn{4}{Ic}{} & {\begin{sideways}\textcolor{blue}{ \texttt Criteria}\end{sideways}} & {\begin{sideways}\textcolor{blue}{ \texttt Fluency}\end{sideways}} & {\begin{sideways}\textcolor{blue}{ \texttt Grammaticality}\end{sideways}} & {\begin{sideways}\textcolor{blue}{ \texttt Correctness}\end{sideways}} & {\begin{sideways}\textcolor{blue}{ \texttt Readability}\end{sideways}} & {\begin{sideways}\textcolor{blue}{ \texttt Understandable}\end{sideways}} &  {\begin{sideways}\textcolor{blue}{ \texttt Relevance}\end{sideways}} & {\begin{sideways}\textcolor{blue}{ \texttt Coherence}\end{sideways}} & {\begin{sideways}\textcolor{blue}{ \texttt Consistency}\end{sideways}} & {\begin{sideways}\textcolor{blue}{ \texttt Sensibleness}\end{sideways}} & {\begin{sideways}\textcolor{blue}{ \texttt Listening}\end{sideways}} & {\begin{sideways}\textcolor{blue}{ \texttt Maintain Context}\end{sideways}} & {\begin{sideways}\textcolor{blue}{ \texttt Logic}\end{sideways}} & \begin{sideways}\textcolor{blue}{ \texttt Informativeness}\end{sideways} & {\begin{sideways}\textcolor{blue}{ \texttt Diversity}\end{sideways}} & {\begin{sideways}\textcolor{blue}{ \texttt Specificity}\end{sideways}} & {\begin{sideways}\textcolor{blue}{ \texttt Proactivity}\end{sideways}} & {\begin{sideways}\textcolor{blue}{ \texttt Flexible}\end{sideways}} & {\begin{sideways}\textcolor{blue}{ \texttt Overall Quality}\end{sideways}} & {\begin{sideways}\textcolor{blue}{ \texttt Appropriateness}\end{sideways}} & \begin{sideways}\textcolor{blue}{ \texttt Naturalness}\end{sideways} & {\begin{sideways}\textcolor{blue}{ \texttt Humanness}\end{sideways}} & {\begin{sideways}\textcolor{blue}{ \texttt Adequacy}\end{sideways}} & {\begin{sideways}\textcolor{blue}{ \texttt Engagingness}\end{sideways}} & {\begin{sideways}\textcolor{blue}{ \texttt Interestingness}\end{sideways}} & {\begin{sideways}\textcolor{blue}{ \texttt Empathy}\end{sideways}} & {\begin{sideways}\textcolor{blue}{ \texttt Emotion}\end{sideways}} & {\begin{sideways}\textcolor{blue}{ \texttt Others}\end{sideways}} \\
    \hline
   Paper & {Venue} & \multicolumn{1}{m{3.5em}|}{Citation Number} & {year} & \multicolumn{1}{m{3.5em}I}{Criteria Number}  & \multicolumn{1}{p{0.5em}|}{\color{blue} 34} & \multicolumn{1}{p{0.5em}|}{\color{blue} 8} & \multicolumn{1}{p{0.5em}|}{\color{blue} 3} & \multicolumn{1}{p{0.5em}|}{\color{blue} 2} & \multicolumn{1}{p{0.5em}I}{3} &  \multicolumn{1}{p{0.5em}|}{\color{blue} 29} & \multicolumn{1}{p{0.5em}|}{\color{blue} 18} & \multicolumn{1}{p{0.5em}|}{\color{blue} 11} & \multicolumn{1}{p{0.5em}|}{\color{blue} 6} & \multicolumn{1}{p{0.5em}|}{\color{blue} 2} & \multicolumn{1}{p{0.5em}|}{\color{blue} 1} & \multicolumn{1}{p{0.5em}I}{1} & \multicolumn{1}{p{0.5em}|}{\color{blue} 26} & \multicolumn{1}{p{0.5em}|}{\color{blue} 14} & \multicolumn{1}{p{0.5em}|}{\color{blue} 3} & \multicolumn{1}{p{0.5em}|}{\color{blue} 2} & \multicolumn{1}{p{0.5em}I}{1} & \multicolumn{1}{p{0.5em}|}{\color{blue} 40} & \multicolumn{1}{p{0.5em}|}{\color{blue} 12} & \multicolumn{1}{p{0.5em}|}{\color{blue} 7} & \multicolumn{1}{p{0.5em}|}{\color{blue} 5} & \multicolumn{1}{p{0.5em}I}{1} & \multicolumn{1}{p{0.5em}|}{\color{blue} 11} & \multicolumn{1}{p{0.5em}I}{9} & \multicolumn{1}{p{0.5em}|}{\color{blue} 6} & \multicolumn{1}{p{0.5em}I}{3} & \multicolumn{1}{p{0.5em}I}{4} \\
    \hline
    \endhead 
    \hline \multicolumn{32}{r}{{Continued on next page}} \\
\endfoot
\endlastfoot
    \citet{li2020don} & ACL& 16    & 2020  & 1            &       &       &       &       &       &       &       &       &       &       &       &       & &       &       &       &       &       &       &  &       &       & {\color{blue}\checkmark}&       &       &       &  \\
    \hline
    \citet{bao2020plato} & ACL & 16    & 2020  & 4     & {{\color{blue}\checkmark}} &       &       &       &              &       & {{\color{blue}\checkmark}} &       &       &       &       &       & {\color{blue}\checkmark}     &       &       &       &       & {{\color{blue}\checkmark}} &       &  &       &       &       &       &       &       &  \\
    \hline
    \citet{shen2020cdl} & ACL & 7     & 2020  & 2     &       &       &       &       &       &       &             &       &       &       &       &       &  &       &       &       &       & {{\color{blue}\checkmark}} &       &  &       &       &       &       &       & {{\color{blue}\checkmark}} &  \\
    \hline
    \citet{mehri2020usr} & ACL & 5     & 2020  & 6     &       &       &       &       & {{\color{blue}\checkmark}} &        {{\color{blue}\checkmark}} &       &       &       &       & {{\color{blue}\checkmark}} &       &  &       &       &       &       & {{\color{blue}\checkmark}} &       & {\color{blue}\checkmark}     &       &       &       & {{\color{blue}\checkmark}} &       &       &  \\
    \hline
    \citet{wu2020diverse} & ACL & 5     & 2020  & 2     &       &       &       &       &       &       &              &       &       &       &       &       & {\color{blue}\checkmark}     &       &       &       &       &       & {{\color{blue}\checkmark}} &  &       &       &       &       &       &       &  \\
    \hline
    \citet{su2020diversifying} & ACL & 4     & 2020  & 3     & {{\color{blue}\checkmark}} &       &       &       &       &       & {{\color{blue}\checkmark}} &       &       &       &             &       &  &       &       &       &       &       &       &  &       &       &       & {\color{blue}\checkmark} &       &       &  \\
    \hline
    \citet{pang2020towards} & ACL & 3     & 2020  & 4     & {{\color{blue}\checkmark}} &       &       &       &       &       &       & {{\color{blue}\checkmark}} & {{\color{blue}\checkmark}} &       &       &       &       &  & {{\color{blue}\checkmark}} &       &       &       &       &       &  &       &       &       &       &              &  \\
    \hline
    \citet{cai2020data} & ACL & 2     & 2020  & 1     &       &       &       &       &       &       &       &       &       &       &       &       &       &  &       &       &       &       & {{\color{blue}\checkmark}} &       &  &             &       &       &       &       &  \\
    \hline
    \citet{song2020generate} & ACL & 1     & 2020  & 4     & {{\color{blue}\checkmark}} &       &       &       &       &       & {{\color{blue}\checkmark}} &       & {{\color{blue}\checkmark}} &       &       &       &       & {\color{blue}\checkmark}     &       &       &       &       &       &       &  &       &       &              &       &       &  \\
    \hline
    \citet{song2020learning} & ACL & 0     & 2020  & 2     &       &       &       &       &       &       &       &       & {{\color{blue}\checkmark}} &       &       &       &       &  &       &       &       &       &       & {{\color{blue}\checkmark}} &  &       &       &       &       &       &         \\
    \hline
    \citet{bak2020speaker} & ACL & 0     & 2020  & 2     &       &       &       &       &       &       & {{\color{blue}\checkmark}} &       &       &       &       &       &       &  &       &       &       &       &       & {{\color{blue}\checkmark}} &  &       &       &       &       &       &         \\
    \hline
    \citet{lin2020generating} & ACL & 0     & 2020  & 3     & {{\color{blue}\checkmark}} &       &       &       &       &       &       & {{\color{blue}\checkmark}} &       &       &       &       &       & {\color{blue}\checkmark}     &       &       &       &       &       &       &  &       &       &       &              &       &  \\
    \hline
    \citet{sinha2020learning} & ACL & 5     & 2020  & 7     & {{\color{blue}\checkmark}} &       &       &       &       &       &       &       &       & {{\color{blue}\checkmark}} & {{\color{blue}\checkmark}} &       &       &  & {{\color{blue}\checkmark}} &       &       &              &       &  & {{\color{blue}\checkmark}} &       & {{\color{blue}\checkmark}} & {{\color{blue}\checkmark}} &       &       & {{\color{blue}\checkmark}} \\
    \hline
    \citet{shum2020sketch} & ACL & 1     & 2020  & 3     & {{\color{blue}\checkmark}} &       &       &       &       &       &       &       &       & {{\color{blue}\checkmark}} &       &       &       &  &       &       &       &       &       &       &  &       &       & {{\color{blue}\checkmark}} &       &              &  \\
    \hline
    \citet{yuma2020ubleu} & ACL & 0     & 2020  & 1     &       &       &       &       &       &       &       &       &       &       &       &       &       &  &       &       &       &       &       & {{\color{blue}\checkmark}} &        &       &       &       &       &       &  \\
    \hline
    \citet{zhao2020learning} & EMNLP & 2     & 2020  & 3     & {{\color{blue}\checkmark}} &       &       &       &       &       &       &       & {{\color{blue}\checkmark}} &       &       &       &       & {\color{blue}\checkmark}     &       &       &             &       &       &  &       &       &       &       &       &       &  \\
    \hline
    \citet{zhao2020knowledge} & EMNLP & 1     & 2020  & 3     & {{\color{blue}\checkmark}} &       &       &              &       & {{\color{blue}\checkmark}} & {{\color{blue}\checkmark}} &       &       &       &       &       &  &       &       &       &       &       &       &  &       &       &       &       &       &       &  \\
    \hline
    \citet{chen2020bridging} & EMNLP & 1     & 2020  & 2     &       &       &       &       &       &       &       &       &       & {{\color{blue}\checkmark}} &       &       &       &  &       & {{\color{blue}\checkmark}} &       &       &       &       &  &       &       &       &       &       &       \\
    \hline
    \citet{ji2020cross} & EMNLP & 0     & 2020  & 2     & {{\color{blue}\checkmark}} &       &       &       &       &       & {{\color{blue}\checkmark}} &       &       &       &       &       &       &  &       &       &       &       &       &       &  &       &       &       &       &       &       \\
    \hline
    \citet{jaques2020human} & EMNLP & 0     & 2020  & 5     & {{\color{blue}\checkmark}} &       &       &       &       &       & {{\color{blue}\checkmark}} &       &       &       &       &       &       &  & {{\color{blue}\checkmark}} &       &       &       & {{\color{blue}\checkmark}} &       &  &       &       &       &       & {{\color{blue}\checkmark}} &       \\
    \hline
    \citet{deriu2020spot} & EMNLP & 0     & 2020  & 3     & {{\color{blue}\checkmark}} &       &       &       &       &       &       &       &       & {{\color{blue}\checkmark}} &       &       &       &  &       & {{\color{blue}\checkmark}} &       &       &       &       &  &       &       &       &       &       &       \\
    \hline
    \citet{wu2020improving} & EMNLP & 0     & 2020  & 2     &       &       &       &       &       &       &       &       &       &       &       &       &       & {\color{blue}\checkmark}     &       &       &       &       &       & {{\color{blue}\checkmark}} &  &       &       &       &       &       &       \\
    \hline
    \citet{ko2020generating} & EMNLP & 0     & 2020  & 3     &       &       &       &       &       &       &       &       &       &       &       &       &       & {\color{blue}\checkmark}     & {{\color{blue}\checkmark}} & {{\color{blue}\checkmark}} &       &       &       &       &  &       &       &       &       &       &       \\
    \hline
    \citet{feng2020regularizing} & EMNLP & 0     & 2020  & 3     & {{\color{blue}\checkmark}} &       &       &       &       &       & {{\color{blue}\checkmark}} &       &       &       &       &       &       &  & {{\color{blue}\checkmark}} &       &       &       &       &       &  &       &       &       &       &       &       \\
    \hline
    \citet{kim2020will} & EMNLP & 0     & 2020  & 2     &       &       &       &       &       &       &       &       & {{\color{blue}\checkmark}} &       &       &       &       &  &       &       &       &       &       &       &  &       &       & {{\color{blue}\checkmark}} &       &       &       \\
    \hline
    \citet{gao2020dialogue} & EMNLP & 0     & 2020  & 4     & {{\color{blue}\checkmark}} &       &       &       &       &       & {{\color{blue}\checkmark}} & {{\color{blue}\checkmark}} &       &       &       &       &       & {\color{blue}\checkmark}     &       &       &       &       &       &       &  &       &       &       &       &       &       \\
    \hline
    \citet{cai2020group} & EMNLP & 0     & 2020  & 4     & {{\color{blue}\checkmark}} &       &       &       &       &       &       & {{\color{blue}\checkmark}} &       &       &       &       &       & {\color{blue}\checkmark}     &       &       &       &       &       &       &  &       &       & {{\color{blue}\checkmark}} &       &       &       \\
    \hline
    \citet{yang2020styledgpt} & EMNLP & 0     & 2020  & 4     & {{\color{blue}\checkmark}} &       &       &       &       &       & {{\color{blue}\checkmark}} &       & {{\color{blue}\checkmark}} &       &       &       &       & {\color{blue}\checkmark}     &       &       &       &       &       &       &  &       &       &       &       &       &       \\
    \hline
    \citet{kong2020tsdg} & EMNLP & 0     & 2020  & 3     & {{\color{blue}\checkmark}} &       &       &       &       &       & {{\color{blue}\checkmark}} &       &       &       &       &       &       & {\color{blue}\checkmark}     &       &       &       &       &       &       &  &       &       &       &       &       &       \\
    \hline
    \citet{takayama2020consistent} & EMNLP & 0     & 2020  & 1     &       &       &       &       &       &       &       &       &       &       &       &       &       &  &       &       &       &       & {{\color{blue}\checkmark}} &       &  &       &       &       &       &       &       \\
    \hline
    \citet{cui2020focus} & EMNLP & 0     & 2020  & 2     &       &       &       &       &       &       &       &       &       &       &       &       &       &  & {{\color{blue}\checkmark}} &       &       &       & {{\color{blue}\checkmark}} &       &  &       &       &       &       &       &       \\
    \hline
    \citet{santhanam2020learning} & EMNLP & 0     & 2020  & 3     &       &       &       &       &       &       &       &       &       &       &       &       &       & {\color{blue}\checkmark}     &       &       &       &       & {{\color{blue}\checkmark}} & {{\color{blue}\checkmark}} &  &       &       &       &       &       &       \\
    \hline
    \citet{zhang2020topic} & COLING & 0     & 2020  & 2     &       & {{\color{blue}\checkmark}} &       &       &       &       & {{\color{blue}\checkmark}} &       &       &       &       &       &       &  &       &       &       &       &       &       &  &       &       &       &       &       &       \\
    \hline
    \citet{ueyama2020diverse} & COLING & 0     & 2020  & 2     &       &       &       &       &       &       &       & {{\color{blue}\checkmark}} &       &       &       &       &       &  &       &       &       &       &       &       &  &       &       & {{\color{blue}\checkmark}} &       &       &       \\
    \hline
    \citet{phy2020deconstruct} & COLING & 0     & 2020  & 4     &       &       &       &       & {{\color{blue}\checkmark}} &       &       &       &       & {{\color{blue}\checkmark}} &       &       &       & {\color{blue}\checkmark}     &       &       &       &       & {{\color{blue}\checkmark}} &       &  &       &       &       &       &       &       \\
    \hline
    \citet{li2020empdg} & COLING & 0     & 2020  & 3     & {{\color{blue}\checkmark}} &       &       &       &       &       & {{\color{blue}\checkmark}} &       &       &       &       &       &       &  &       &       &       &       &       &       &  &       &       &       &       & {{\color{blue}\checkmark}} &       \\
    \hline
    \citet{mehri2020unsupervised} & SIGDIAL & 5     & 2020  & 13    & {{\color{blue}\checkmark}} &       & {{\color{blue}\checkmark}} &       & {{\color{blue}\checkmark}} &       & {{\color{blue}\checkmark}} & {{\color{blue}\checkmark}} & {{\color{blue}\checkmark}} &       &       &       &       & {\color{blue}\checkmark}     & {{\color{blue}\checkmark}} &       &       & {{\color{blue}\checkmark}} & {{\color{blue}\checkmark}} & {{\color{blue}\checkmark}} &  &       &        {{\color{blue}\checkmark}} & {{\color{blue}\checkmark}} &       &       & {{\color{blue}\checkmark}} \\
    \hline
    \citet{finch2020towards} & SIGDIAL& 1     & 2020  & 8     &       & {{\color{blue}\checkmark}} &       &       &       &       & {{\color{blue}\checkmark}} &       & {{\color{blue}\checkmark}} &       &       &       &       & {\color{blue}\checkmark}     &       &       & {{\color{blue}\checkmark}} &       & {{\color{blue}\checkmark}} &       &  &       &       & {{\color{blue}\checkmark}} &       &       & {{\color{blue}\checkmark}} \\
    \hline
    \citet{reed2020learning} & SIGDIAL & 1     & 2020  & 3     &       & {{\color{blue}\checkmark}} &       &       &       &       &       & {{\color{blue}\checkmark} } &       &       &       &       &       &  &       &       &       &       &       &       & {\color{blue}\checkmark}     &       &       &       &       &       &       \\
    \hline
    \citet{hsueh2020semantic} & SIGDIAL & 0     & 2020  & 1     &       &       &       &       &       &       &       &       &       &       &       &       &       &  &       &       &       &       & {{\color{blue}\checkmark}} &       &  &       &       &       &       &       &       \\
    \hline
    \citet{hedayatnia2020policy} & INLG & 3     & 2020  & 1     &       &       &       &       &       &       &       &       &       &       &       &       &       &  &       &       &       &       &       & {{\color{blue}\checkmark}} &  &       &       &       &       &       &       \\
    \hline
    \citet{rashkin2019towards} & ACL & 80    & 2019  & 3     & {{\color{blue}\checkmark}} &       &       &       &       &       & {{\color{blue}\checkmark}} &       &       &       &       &       &       &  &       &       &       &       &       &       &  &       &       &       &       & {{\color{blue}\checkmark}} &       \\
    \hline
    \citet{madotto2019personalizing} & ACL & 50    & 2019  & 2     & {{\color{blue}\checkmark}} &       &       &       &       &       &       &       & {{\color{blue}\checkmark}} &       &       &       &       &  &       &       &       &       &       &       &  &       &       &       &       &       &       \\
    \hline
    \citet{wu2019proactive} & ACL & 32    & 2019  & 4     & {{\color{blue}\checkmark}} &       &       &       &       &       &       & {{\color{blue}\checkmark}} &       &       &       &       &       & {\color{blue}\checkmark}     &       &       & {{\color{blue}\checkmark}} &       &       &       &  &       &       &       &       &       &       \\
    \hline
    \citet{li2019incremental} & ACL & 32    & 2019  & 3     & {{\color{blue}\checkmark}} &       &       &       &       &       & {{\color{blue}\checkmark}} & {{\color{blue}\checkmark}} &       &       &       &       &       &  &       &       &       &       &       &       &  &       &       &       &       &       &       \\
    \hline
    \citet{dziri2019augmenting} & ACL & 27    & 2019  & 1     &       &       &       &       &       &       &       &       &       &       &       &       &       &  &       &       &       &       & {{\color{blue}\checkmark}} &       &  &       &       &       &       &       &       \\
    \hline
    \citet{su2019improving} & ACL & 19    & 2019  & 1     & {{\color{blue}\checkmark}} &       &       &       &       &       &       &       &       &       &       &       &       &  &       &       &       &       &       &       &  &       &       &       &       &       &       \\
    \hline
    \citet{qiu2019training} & ACL & 19    & 2019  & 2     &       &       &       & {{\color{blue}\checkmark}} &       &       &       &       &       &       &       &       &       &  & {{\color{blue}\checkmark}} &       &       &       &       &       &  &       &       &       &       &       &       \\
    \hline
    \citet{zhu2019retrieval} & ACL & 14    & 2019  & 3     &       & {{\color{blue}\checkmark}} &       &       &       &       &       &       &       &       &       &       &       & {\color{blue}\checkmark}     &       &       &       &       &       & {{\color{blue}\checkmark}} &  &       &       &       &       &       &       \\
    \hline
    \citet{xu2019neural} & ACL & 13    & 2019  & 1     &       &       &       &       &       &       &       &       &       &       &       &       &       &  &       &       &       &       & {{\color{blue}\checkmark}} &       &  &       &       &       &       &       &       \\
    \hline
    \citet{tian2019learning} & ACL & 12    & 2019  & 2     &       &       &       &       &       &       &       &       &       &       &       &       &       & {\color{blue}\checkmark}     &       &       &       &       & {{\color{blue}\checkmark}} &       &  &       &       &       &       &       &       \\
    \hline
    \citet{bao2019know} & ACL & 4     & 2019  & 4     &       &       &       &       &       &       &       & {{\color{blue}\checkmark}} &       &       &       &       &       & {\color{blue}\checkmark}     & {{\color{blue}\checkmark}} &       &       &       & {{\color{blue}\checkmark}} &       &  &       &       &       &       &       &       \\
    \hline
    \citet{zhang2020dialogpt} & ACL & 119   & 2019  & 3     &       &       &       &       &       &       & {{\color{blue}\checkmark}} &       &       &       &       &       &       & {\color{blue}\checkmark}     &       &       &       &       &       &       &  & {{\color{blue}\checkmark}} &       &       &       &       &       \\
    \hline
    \citet{olabiyi2019multi} & ACL & 19    & 2019  & 1     &       &       &       &       &       &       &       &       &       &       &       &       &       &  &       &       &       &       & {{\color{blue}\checkmark}} &       &  &       &       &       &       &       &       \\
    \hline
    \citet{lin2019moel} & EMNLP & 22    & 2019  & 3     & {{\color{blue}\checkmark}} &       &       &       &       &       & {{\color{blue}\checkmark}} &       &       &       &       &       &       &  &       &       &       &       &       &       &  &       &       &       &       & {{\color{blue}\checkmark}} &       \\
    \hline
    \citet{gao2019structuring} & EMNLP & 14    & 2019  & 1     &       &       &       &       &       &       &       &       &       &       &       &       &       &  &       &       &       &       &       & {{\color{blue}\checkmark}} &  &       &       &       &       &       &       \\
    \hline
    \citet{qin2019entity} & EMNLP & 13    & 2019  & 3     & {{\color{blue}\checkmark}} &       & {{\color{blue}\checkmark}} &       &       &       &       &       &       &       &       &       &       &  &       &       &       &       &       &       &  & {{\color{blue}\checkmark}} &       &       &       &       &       \\
    \hline
    \citet{gao2019structuring} & EMNLP & 10    & 2019  & 1     &       &       &       &       &       &       &       &       &       &       &       &       &       &  &       &       &       &       & {{\color{blue}\checkmark}} &       &  &       &       &       &       &       &       \\
    \hline
    \citet{pan2019improving} & EMNLP & 9     & 2019  & 2     & {{\color{blue}\checkmark}} &       &       &       &       &       &       &       &       &       &       &       &       &  &       &       &       &       & {{\color{blue}\checkmark}} &       &  &       &       &       &       &       &       \\
    \hline
    \citet{zeng2019dirichlet} & EMNLP & 6     & 2019  & 1     &       &       &       &       &       &       &       &       &       &       &       &       &       &  &       &       &       &       & {{\color{blue}\checkmark}} &       &  &       &       &       &       &       &       \\
    \hline
    \citet{zhou2019unsupervised} & EMNLP & 5     & 2019  & 1     &       &       &       &       &       &       &       &       &       &       &       &       &       &  &       &       &       &       & {{\color{blue}\checkmark}} &       &  &       &       &       &       &       &       \\
    \hline
    \citet{yang2019low} & EMNLP & 3     & 2019  & 3     & {{\color{blue}\checkmark}} &       &       &       &       &       & {{\color{blue}\checkmark}} &       &       &       &       &       &       & {\color{blue}\checkmark}     &       &       &       &       &       &       &  &       &       &       &       &       &       \\
    \hline
    \citet{chang2019semi} & EMNLP & 2     & 2019  & 1     &       &       &       &       &       &       &       &       &       &       &       &       &       &  &       &       &       &       & {{\color{blue}\checkmark}} &       &  &       &       &       &       &       &       \\
    \hline
    \citet{see2019makes} & NAACL & 59    & 2019  & 7     & {{\color{blue}\checkmark}} &       &       &       &       &       &       &       &       & {{\color{blue}\checkmark}} & {{\color{blue}\checkmark}} &       &       &  & {{\color{blue}\checkmark}} &       &       &       &       &       &  & {{\color{blue}\checkmark}} &       & {{\color{blue}\checkmark}} & {{\color{blue}\checkmark}} &       &        {{\color{blue}\checkmark}} \\
    \hline
    \citet{gao2019jointly} & NAACL & 42    & 2019  & 2     &       &       &       &       &       &       & {{\color{blue}\checkmark}} &       &       &       &       &       &       &  &       &       &       &       &       &       &  &       &       &       & {{\color{blue}\checkmark}} &       &       \\
    \hline
    \citet{cai2019skeleton} & NAACL & 17    & 2019  & 1     &       &       &       &       &       &       &       &       &       &       &       &       &       &  &       &       &       &       & {{\color{blue}\checkmark}} &       &  &       &       &       &       &       &       \\
    \hline
    \citet{ko2019linguistically} & NAACL & 12    & 2019  & 3     &       &       &       &       &       &       & {{\color{blue}\checkmark}} &       &       &       &       &       &       & {\color{blue}\checkmark}     &       &       &       &       &       &       & {\color{blue}\checkmark}     &       &       &       &       &       &       \\
    \hline
    \citet{sedoc2019chateval} & NAACL & 10    & 2019  & 1     &       &       &       &       &       &       &       &       &       &       &       &       &       &  &       &       &       &       & {{\color{blue}\checkmark}} &       &  &       &       &       &       &       &       \\
    \hline
    \citet{ghazarian2019better} & NAACL & 16    & 2019  & 1     &       &       &       &       &       &       &       &       &       &       &       &       &       &  &       &       &       &       &       & {{\color{blue}\checkmark}} &  &       &       &       &       &       &       \\
    \hline
    \citet{cui2019dal} & NAACL & 6     & 2019  & 1     &       &       &       &       &       &       &       &       &       &       &       &       &       &  &       &       &       &       & {{\color{blue}\checkmark}} &       &  &       &       &       &       &       &       \\
    \hline
    \citet{sankar2019deep} & SIGDIAL & 14    & 2019  & 2     &       &       &       &       &       &       & {{\color{blue}\checkmark}} &       &       &       &       &       &       &  & {{\color{blue}\checkmark}} &       &       &       &       &       &  &       &       &       &       &       &       \\
    \hline
    \citet{kulikov2019importance} & INLG & 17    & 2019  & 1     &       &       &       &       &       &       &       &       &       &       &       &       &       &  &       &       &       &       & {{\color{blue}\checkmark}} &       &  &       &       &       &       &       &       \\
    \hline
    \citet{yi2019towards} & INLG & 13    & 2019  & 2     &       &       &       &       &       &       &       & {{\color{blue}\checkmark}} &       &       &       &       &       &  &       &       &       &       &       &       &  &       &       & {{\color{blue}\checkmark}} &       &       &       \\
    \hline
    \citet{santhanam2019towards} & INLG & 5     & 2019  & 2     &       &       &       & {{\color{blue}\checkmark}} &       &       &       & {{\color{blue}\checkmark}} &       &       &       &       &       &  &       &       &       &       &       &       &  &       &       &       &       &       &       \\
    \hline
    \citet{deriu2019towards} & INLG & 1     & 2019  & 1     &       &       &       &       &       &       &       &       &       &       &       &       &       &  &       &       &       &       & {{\color{blue}\checkmark}} &       &  &       &       &       &       &       &       \\
    \hline
    \citet{liu2018knowledge} & ACL & 80    & 2018  & 3     & {{\color{blue}\checkmark}} &       & {{\color{blue}\checkmark}} &       &       &       & {{\color{blue}\checkmark}} &       &       &       &       &       &       &  &       &       &       &       &       &       &  &       &       &       &       &       &       \\
    \hline

    \citet{zhang2018learning} & ACL & 52    & 2018  & 1     &       &       &       &       &       &       &       &       &       &       &       &       &       &  &       &       &       &       & {{\color{blue}\checkmark}} &       &  &       &       &       &       &       &       \\
    \hline
    \citet{ke2018generating} & ACL & 36    & 2018  & 3     &       & {{\color{blue}\checkmark}} &       &       &       &       &       &       &       &       &       &       &       & {\color{blue}\checkmark}     &       &       &       &       &       & {{\color{blue}\checkmark}} &  &       &       &       &       &       &       \\
    \hline
    \citet{moghe2018towards} & EMNLP & 53    & 2018  & 3     & {{\color{blue}\checkmark}} &       &       &       &       &       & {{\color{blue}\checkmark}} &       &       &       &       &       &       &  &       &       &       &       &       &       &  & {{\color{blue}\checkmark}} &       &       &       &       &       \\
    \hline
    \citet{baheti2018generating} & EMNLP & 39    & 2018  & 2     &       &       &       &       &       &       &       &       &       &       &       &       &       & {\color{blue}\checkmark}     &       &       &       &       &       &       & {\color{blue}\checkmark}     &       &       &       &       &       &       \\
    \hline
    \citet{du2018variational} & EMNLP & 25    & 2018  & 3     &       & {{\color{blue}\checkmark}} &       &       &       &       & {{\color{blue}\checkmark}} & {{\color{blue}\checkmark}} &       &       &       &       &       &  &       &       &       &       &       &       &  &       &       &       &       &       &       \\
    \hline
    \citet{luo2018auto} & EMNLP & 22    & 2018  & 2     & {{\color{blue}\checkmark}} &       &       &       &       &       &       & {{\color{blue}\checkmark}} &       &       &       &       &       &  &       &       &       &       &       &       &  &       &       &       &       &       &       \\
    \hline
    \citet{li2018syntactically} & EMNLP & 18    & 2018  & 3     &       &       &       &       &       &       &       &       & {{\color{blue}\checkmark}} &       &       &       & {{\color{blue}\checkmark}} &  &       &       &       &       &       &       &  &       &       &       &       &       & {{\color{blue}\checkmark}} \\
    \hline
    \citet{liu2018towards} & EMNLP & 16    & 2018  & 2     & {{\color{blue}\checkmark}} &       &       &       &       &       & {{\color{blue}\checkmark}} &       &       &       &       &       &       &  &       &       &       &       &       &       &  &       &       &       &       &       &       \\
    \hline
    \citet{pei2018s2spmn} & EMNLP & 11    & 2018  & 1     &       &       &       &       &       &       &       &       &       &       &       &       &       &  &       &       &       &       & {{\color{blue}\checkmark}} &       &  &       &       &       &       &       &       \\
    \hline
    \citet{shen2018nexus} & EMNLP & 8     & 2018  & 3     & {{\color{blue}\checkmark}} &       &       &       &       &       &       &       & {{\color{blue}\checkmark}} &       &       &       &       &  &       &       &       &       &       &       &  &       &       &       & {{\color{blue}\checkmark}} &       &       \\
    \hline

    \citet{zhang2018context} & COLING & 19    & 2018  & 2     &       &       &       &       &       &       &       & {{\color{blue}\checkmark}} &       &       &       &       &       &  &       &       &       &       &       &       & {\color{blue}\checkmark}     &       &       &       &       &       &       \\
    \hline
    \citet{wu2018hl} & COLING & 2     & 2018  & 1     &       &       &       &       &       &       &       &       &       &       &       &       &       &  &       &       &       &       & {{\color{blue}\checkmark}} &       &  &       &       &       &       &       &       \\
    \hline
    
    \citet{zou2018memd} & COLING & 2     & 2018  & 1     &       &       &       &       &       &       &       &       &       &       &       &       &       &  &       &       &       &       & {{\color{blue}\checkmark}} &       &  &       &       &       &       &       &       \\
    \hline
    \citet{wang2018prospective} & COLING & 1     & 2018  & 2     &       & {{\color{blue}\checkmark}} &       &       &       &       & {{\color{blue}\checkmark}} &       &       &       &       &       &       &  &       &       &       &       &       &       &  &       &       &       &       &       &       \\
    \hline
    \citet{wu2018dialog} & NAACL & 12    & 2018  & 1     &       &       &       &       &       &       &       &       &       &       &       &       &       &  &       &       &       &       & {{\color{blue}\checkmark}} &       &  &       &       &       &       &       &       \\
    \hline
    \citet{xu2018lsdscc} & NAACL & 9     & 2018  & 2     &       &       &       &       &       &       & {{\color{blue}\checkmark}} &       &       &       &       &       &       &  & {{\color{blue}\checkmark}} &       &       &       &       &       &  &       &       &       &       &       &       \\
    \hline
    \citet{lowe2017towards} & ACL & 207   & 2017  & 1     &       &       &       &       &       &       &       &       &       &       &       &       &       &  &       &       &       &       & {{\color{blue}\checkmark}} &       &  &       &       &       &       &       &       \\
    \hline
    \citet{shen2017conditional} & ACL & 94    & 2017  & 3     &       & {{\color{blue}\checkmark}} &       &       &       &       &       & {{\color{blue}\checkmark}} &       &       &       &       &       &  & {{\color{blue}\checkmark}} &       &       &       &       &       &  &       &       &       &       &       &       \\
    \hline
    \citet{li2017adversarial} & EMNLP & 663   & 2017  & 1     &       &       &       &       &       &       &       &       &       &       &       &       &       &  &       &       &       &       & {{\color{blue}\checkmark}} &       &  &       &       &       &       &       &       \\
    \hline
    \citet{shao2017generating} & EMNLP & 136   & 2017  & 1     &       &       &       &       &       &       &       &       &       &       &       &       &       &  &       &       &       &       & {{\color{blue}\checkmark}} &       &  &       &       &       &       &       &       \\
    \hline
    \citet{yao2017towards} & EMNLP & 63    & 2017  & 1     &       &       &       &       &       &       &       &       &       &       &       &       &       &  &       &       &       &       & {{\color{blue}\checkmark}} &       &  &       &       &       &       &       &       \\
    \hline
    \citet{xu2017neural} & EMNLP & 62    & 2017  & 1     &       &       &       &       &       &       &       &       &       &       &       &       &       &  & {{\color{blue}\checkmark}} &       &       &       &       &       &  &       &       &       &       &       &       \\
    \hline
    \citet{wang2017steering} & EMNLP & 57    & 2017  & 1     &       &       &       &       &       &       &       &       &       &       &       &       &       &  &       &       &       &       & {{\color{blue}\checkmark}} &       &  &       &       &       &       &       &       \\
    \hline
    \citet{li2016deep} & EMNLP & 818   & 2016  & 1     &       &       &       &       &       &       &       &       &       &       &       &       &       &  &       &       &       &       & {{\color{blue}\checkmark}} &       &  &       &       &       &       &       &       \\
    \hline
    \citet{liu2016not} & EMNLP & 754   & 2016  & 1     &       &       &       &       &       &       &       &       &       &       &       &       &       &  &       &       &       &       &       &       &  &       & {{\color{blue}\checkmark}} &       &       &       &       \\
    \hline
    \citet{vougiouklis2016neural} & COLING & 28    & 2016  & 1     &       &       &       &       &       &       &       &       &       &       &       &       &       &  &       &       &       &       & {{\color{blue}\checkmark}} &       &  &       &       &       &       &       &       \\
    \hline
    \citet{li2016diversity} & NAACL & 1003  & 2016  & 1     &       &       &       &       &       &       &       &       &       &       &       &       &       &  &       &       &       &       & {{\color{blue}\checkmark}} &       &  &       &       &       &       &       &       \\
    \hline
    \citet{wen2016multi} & NAACL & 133   & 2016  & 2     &       &       &       &       &       &       &       &       &       &       &       &       &       & {\color{blue}\checkmark}     &       &       &       &       &       &       & {\color{blue}\checkmark}     &       &       &       &       &       &       \\
    \hline

\end{longtable}%
\end{landscape}

of papers employ criteria without clear definitions~\cite{song2020generate,lin2020generating,zhao2020learning}. These problems make it not easy to decide the definitions of criteria from chatbot papers. Thus, towards exact definitions, we also conduct investigation in online dictionaries, education papers, linguistic papers, and other online materials. \par
\begin{figure}[t]
    \centering
    \includegraphics[width=0.88\textwidth]{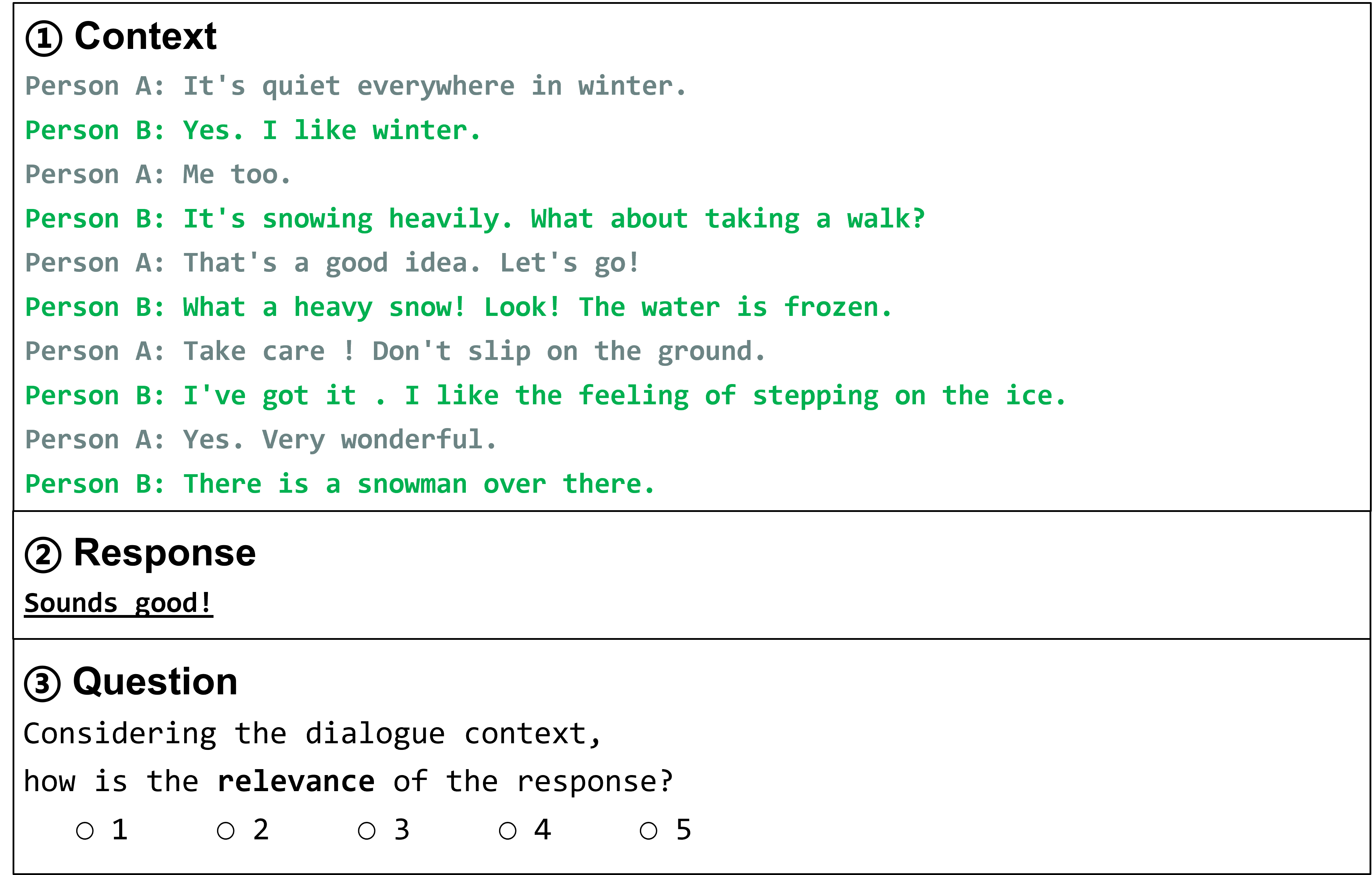}
    \caption{A typical setting of static human evaluation for chatbots.
    {\large\textcircled{\normalsize 1}} Context: a multi-turn dialogue context between two people~(\textcolor{ForestGreen}{\bf Person A} and \textcolor{gray}{\bf Person B}). 
    {\large\textcircled{\normalsize 2}} Response: an utterance answering the last speaking person.
    {\large\textcircled{\normalsize 3}} Question: a set of questions on the response.}
    \label{fig:demo}
\end{figure}

It also worth mentioning that there are three kind of strategies to involve human trials into the evaluation process. The first and the most popular one is static human evaluation --- given a dialogue context between two human and a system generated response, the worker is required to evaluate this response on a criterion, as shown in Figure~\ref{fig:demo}. The second strategy is self-play human evaluation --- let bots chat with each other for multiple turns, the worker is required to evaluate the chatting logs. The last one is interactive human evaluation --- the worker is required to chat with the bot for multiple turns, and evaluation the interaction experience. In this paper, we concentrate on the most commonly used strategy, i.e., the static human evaluation.
\section{Criteria and Definitions}
In this section, we analyze the criteria and definitions in three steps. 1) We present each criterion's definitions in the order of online dictionaries~(\texttt{DIC}), education/linguistic papers+other online materials~(\texttt{ELO})\footnote{As definitions in education/linguistic papers and other online materials are relatively less than dictionaries and chatbot papers), we prefer to analyze education/linguistic papers and other online materials together.}, and chatbot papers~(\texttt{CHAT}). 2)  We summarize the \texttt{DIC}, \texttt{ELO}, and \texttt{CHAT} definitions of this criterion. Such summaries are constructed by the most significant words phrases in source definitions\footnote{If there is only on source definition, we use it as the summary directly.}. We also propose the definition of this criteria based on these summaries. The definition may be the one of the \texttt{DIC}, \texttt{ELO}, and \texttt{CHAT} summaries, or be a rewrote one. The decision is made by more than three experts in chatbots and CHI. 3) At the end of each group, we compare the criteria in this group and conclude our suggestions towards the standard criteria. Note that, the analysis doesn't involve \texttt{Group 6}, as we find that ``\texttt{Empathy}'' and ``\texttt{Emotion}'' are only used for testing emotional chatbots. \texttt{Group 7} is also not involved, as the criteria in this group are barely used.
\subsection{Group 1}
\paragraph{Fluency~(Table~\ref{tab:fluency})} 
\begin{itemize}
    \item \texttt{DIC}: the ability to use a language easily in speaking or writing.
    \item \texttt{ELO}: one’s speaking ability to express the information with ease, minor mistakes are acceptable.
    \item \texttt{CHAT}: the grammatical correctness of generated responses.
    \item \texttt{One’s speaking ability to express the information with ease, minor mistakes are acceptable.}
\end{itemize}
\paragraph{Grammaticality} 
\begin{itemize}
    \item \texttt{DIC}: the quality of conforming to grammar rules.
    \item \texttt{ELO}: the conformity of a sentence to grammar rules.
    \item \texttt{CHAT}: the grammatical well-formedness of a response.
    \item \texttt{The conformity of a response to grammar rules.}
\end{itemize}
\paragraph{Correctness} 
\begin{itemize}
    \item \texttt{DIC}: the conformity to truth.
    \item \texttt{ELO}: the conformity to grammar rules.
    \item \texttt{CHAT}: (Question) Is the response correct or was there a misunderstanding of the conversation?~\cite{mehri2020unsupervised}
    \item \texttt{The conformity of a response to grammar rules.}
\end{itemize}
\paragraph{Readability} 
\begin{itemize}
    \item \texttt{DIC}: the quality of written language to be understood easily.
    \item \texttt{ELO}: the ability of the text to be understood easily.
    \item \texttt{CHAT}: the linguistic quality of text and helps quantify the difficulty of understanding the text by the reader.
    \item \texttt{The quality of the response to be understood easily.}
\end{itemize}
\paragraph{Understandable} 
\begin{itemize}
    \item \texttt{DIC}: the capability of being understood.
    \item \texttt{ELO}: the extent of being understood.
    \item \texttt{CHAT}: N.A.
    \item \texttt{The quality of the response to be understood.}
\end{itemize}
\paragraph{Suggestions} ``\texttt{Grammaticality}'' and ``\texttt{Correctness}'' are the same in definition. Both refer to the conformity of a response to grammar rules. However, it is not necessary employ human to annotate ``\texttt{Grammaticality}'' or ``\texttt{Correctness}, which can be done automatically based on grammar rules. ``Understandable'' and ``Readability'' are very similer in definition. Both are related to the understandability of the response. We think ``Readability'' is  better than ``Grammaticality''. As ``Readability'' is more specific with ``\textit{easily}'', so that workers can do annotations more easily. The definition of ``\texttt{Fluency}'' also indicates ``\textit{easily}''. However, this criterion is not suitable for the static human evaluation, as it describes the ability of the chatbot rather than the quality of a response.  Therefore, though ``\texttt{Fluency}'' is the most frequently used criterion in this group, we prefer to use ``Readability'' as a standard criterion.
\begin{landscape}

\begin{table}[htbp]
\scriptsize
  \centering
  \caption{Details of \texttt{Fluency} w.r.t. source, source type, definition, and question type}
    \begin{tabular}{|l|c|m{60em}|l|}
    \hline
    {Source} & Source Type & Definition & Question Type\\
    \hline
    \hline
    {\href{https://www.thefree Online Dictionary.com/fluency?web=1&wdLOR=c76CF7FF2-F01C-134E-98D0-51EF7E3648E0}{Collins English  Online Dictionary}} &  Online Dictionary & The quality of being able to express oneself readily and effortlessly, esp facility in speech or writing & {}\\
    \hline
    {\href{https://www.thefree Online Dictionary.com/fluency?web=1&wdLOR=c76CF7FF2-F01C-134E-98D0-51EF7E3648E0}{Thesaurus}} &  Online Dictionary & Powerful and effective language & {}\\
    \hline
    {\href{https://www.thefree Online Dictionary.com/fluency?web=1&wdLOR=c76CF7FF2-F01C-134E-98D0-51EF7E3648E0}{Thesaurus}} &  Online Dictionary & skillfulness in speaking or writing & {}\\
    \hline
    {\href{https://www.thefree Online Dictionary.com/fluency?web=1&wdLOR=c76CF7FF2-F01C-134E-98D0-51EF7E3648E0}{Thesaurus}} &  Online Dictionary & The quality of being facile in speech and writing & {}\\
    \hline
    {\href{https://www.thefree Online Dictionary.com/fluency?web=1&wdLOR=c76CF7FF2-F01C-134E-98D0-51EF7E3648E0}{Roget's Thesaurus}} &  Online Dictionary & Ready skill in expression & {}\\
    \hline
    \href{https://www.merriam-webster.com/ Online Dictionary/fluency}{Merriam-Webster} &  Online Dictionary & The quality or state of being capable of using a language easily and accurately & {}\\
    \hline
    \href{https://www.oxfordlearnersdictionaries.com/definition/english/fluency?q=fluency}{Oxford Advanced Learner's  Online Dictionary} &  Online Dictionary & The quality of being able to speak or write a language, especially a foreign language, easily and well & {}\\
    \hline
    \href{https:// Online Dictionary.cambridge.org/ Online Dictionary/english/fluency}{Cambridge  Online Dictionary} &  Online Dictionary & The ability to speak or write a language easily, well, and quickly & {}\\
    \hline
    \hline
    {\citet{kuhn2006teaching}} & Education Paper & The ability of a reader to 1. recognize word quick and accurate, 2. use appropriate prosody, 3. understand and enjoy text & {}\\
    \hline
    {\citet{rasinski2004assessing}} & Education Paper & More than reading fast: reading at an appropriately fast rate with good expression and phrasing that reflects solid understanding of the passage; in second language & {}\\
    \hline
    {\citet{chambers1997we}} & Linguistic Paper & 1. Smooth, rapid, effortless use of language 2. Effectiveness of language use within the constraints of limited linguistic knowledge & {}\\
    \hline
    \href{https://www.readingrockets.org/helping/target/fluency}{Reading Rockets} & Online Materials & The ability to read with speed, accuracy, and proper expression & {}\\
    \hline
    \href{https://www.teachingenglish.org.uk/article/teaching-speaking-unit-9-fluency}{British Council} & Online Materials & Speaking easily, reasonably quickly and without having to stop and pause a lot; how well to communicate meaning rather than how correct to use grammar, pronunciation and vocabulary; not so much about speaking quickly, as communicating the message effectively & {}\\
    \hline
     \href{https://www.teachingenglish.org.uk/sites/teacheng/files/TeachingSpeaking_9_Fuency_v02a\%20\%281\%29.pdf}{British Council} & Online Materials & The ability to communicate meaning without too much stopping or hesitating,  & {}\\
    \hline
    \href{https://study.com/academy/lesson/teaching-strategies-for-reading-writing-fluency.html}{Study.com} & Online Materials & The ability to write with a natural flow and rhythm & {}\\
    \hline
   \href{https://www.etprofessional.com/going-beyond-accuracy-vs-fluency}{English Teaching professional} & Online Materials & Mostly describe the speaking ability, "written fluency": the ability to write with ease and with a flow that is not disrupted by hesitations or the need to search for the right words or phrases to use. Fluent speaker might speak with the odd mistake and that this is accepted & {}\\
    \hline
   
    \href{https://www.bbc.com/future/article/20190903-linguistic-fluency-proficiency-second-language-learning}{BBC Future} & Online Materials & How smoothly and efficiently a second language speaker can speak on a range of topics in real time; While fluency may denote a degree of proficiency, it does not automatically imply accuracy – the ability to produce grammatically correct sentences – nor does it imply grammatical range & {}\\
    \hline\hline
    \citet{bao2020plato}      & Chatbot Paper & Whether the generated sentence is smooth and grammatically correct & Likert rating\\
    \hline
    \citet{pang2020towards}      & Chatbot Paper & (Language Fluency) the quality of phrasing relative to a human native speaker & Likert rating\\
    \hline
    \citet{sinha2020learning}      & Chatbot Paper & Question: How naturally did this user speak English? & Likert rating\\
    \hline
    \citet{shum2020sketch}      & Chatbot Paper & Whether responses are grammatically correct and sound natural & Likert rating\\
    \hline
    \citet{ji2020cross}      & Chatbot Paper & Generate utterance is readablity and grammatical correctness & Likert rating\\
    \hline
    \citet{deriu2019towards}      & Chatbot Paper & Question: Which entities' language is more fluent and grammatically correct?  & Pairwise comparison\\
    \hline
     \citet{feng2020regularizing}     & Chatbot Paper & Question: how likely the generated response is from human? & Pairwise comparison\\
    \hline
     \citet{gao2020dialogue}     & Chatbot Paper & the grammatical correctness of responses & Pairwise comparison\\
    \hline
    \citet{yang2020styledgpt}      & Chatbot Paper & If the response is fluent without any grammatical errors? & Pairwise comparison\\
    \hline
     \citet{phy2020deconstruct}     & Chatbot Paper & The grammatical correctness and readability of the generated responses & Likert rating\\
    \hline
     \citet{mehri2020unsupervised}     & Chatbot Paper & Question: Is the response fluently written? & Multiple choices\\
    \hline
     \citet{rashkin2019towards}     & Chatbot Paper & Question: Could you understand the responses? Did the language seem accurate? & Likert rating\\
    \hline
    \citet{wu2019proactive}      & Chatbot Paper & If the produced response itself is fluent & Likert rating\\
    \hline
     \citet{li2019incremental}     & Chatbot Paper & Whether the response is natural (comprehensible) and fluent & Likert rating\\
    \hline
    \citet{lin2019moel}      & Chatbot Paper & Question: Could you understand the responses from the LISTENER? Did the language seem accurate? & Likert rating\\
    \hline
   \citet{yang2019low}       & Chatbot Paper & If the response is fluent without grammatical error & Pairwise comparison\\
    \hline
   \citet{see2019makes}       & Chatbot Paper & How naturally did this user speak English? & Multiple choices\\
    \hline
    \citet{ghandeharioun2019approximating}      & Chatbot Paper & Question: How FLUENT was the chat bot? (i.e., did it use correct grammar and sentence structure) & Likert rating\\
    
    \hline
     \citet{luo2018auto}     & Chatbot Paper & Whether each sentence is in correct grammar & Likert rating\\
    \hline
   \citet{liu2018towards}      & Chatbot Paper & the readablity and grammatical correctness & Likert rating\\
    \hline
    \citet{shen2018nexus}      & Chatbot Paper & whether the response itself is a fluent natural sentence & Yes-no judgement\\
    \hline
    \end{tabular}%
  \label{tab:fluency}%
\end{table}%
\end{landscape}
\subsection{Group 2}
\paragraph{Relevance}
\begin{itemize}
    \item \texttt{DIC:} the connection of something with the matter at hand.
    \item \texttt{ELO:} the ability to connect with the matter at hand (information selection and organization).
    \item \texttt{CHAT:} the quality of the response being on-topic with the context.
    \item \texttt{The quality of a response to connect with the context.}
\end{itemize}
\paragraph{Coherence}
\begin{itemize}
    \item \texttt{DIC:} the quality of all the parts being logically organized.
    \item \texttt{ELO:}  the quality of all sentences organized logically.
    \item \texttt{CHAT:} N.A.
    \item \texttt{(Local coherence) The quality of a response to connect with the context; (Global Coherence) The quality of all sentences organized logically.}
\end{itemize}
\paragraph{Consistency}
\begin{itemize}
    \item \texttt{DIC:} the quality of agreement among related things.
    \item \texttt{ELO:} the orderly presentation of a set of linked elements in the text.
    \item \texttt{CHAT:} the quality of response agreeing with (not contradicting) the known information (common sense, pre-configured knowledge, etc.), as shown in Figure~\ref{fig:consistency}.
    \item \texttt{The quality of a response agreeing with the known information (common sense, context, etc.).}
    \begin{figure}
        \centering
        \includegraphics[width=0.5\textwidth]{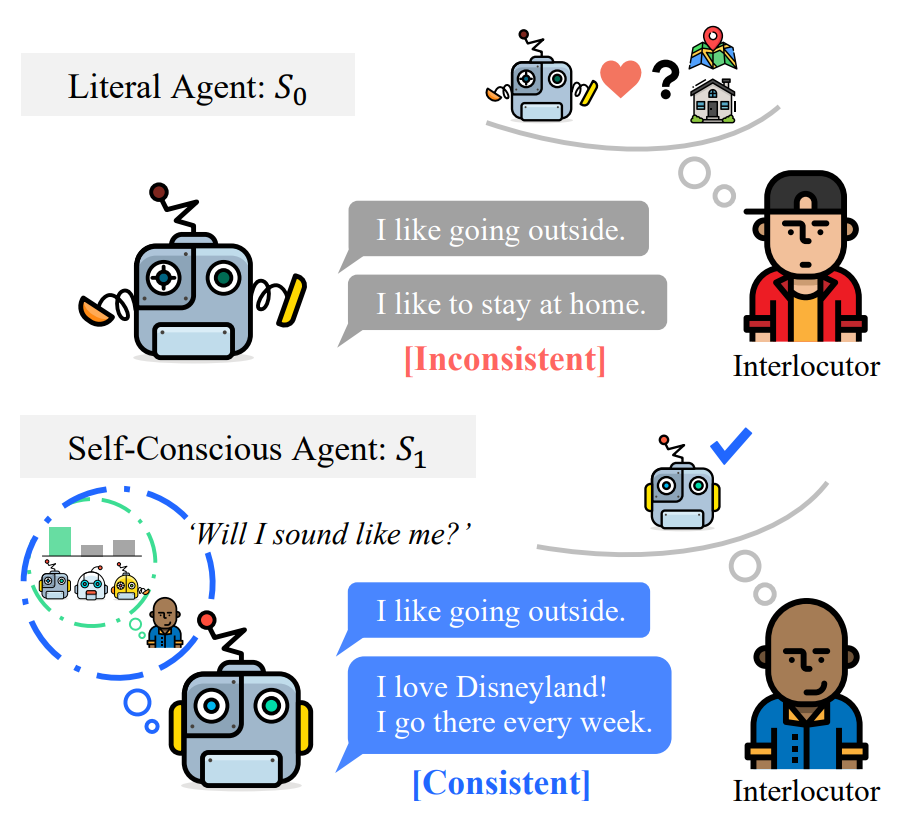}
        \caption{An example of consistency and inconsistency in \cite{kim2020will}}
        \label{fig:consistency}
    \end{figure}
\end{itemize}
\paragraph{Sensibleness}
\begin{itemize}
    \item \texttt{DIC:} the quality of showing good judgement.
    \item \texttt{ELO:} the quality of showing good judgement.
    \item \texttt{CHAT:} the quality of the response being understandable, logically coherent, consistency and conforming to common sense.
    \item \texttt{The quality of the response being understandable, logically coherent, consistency and conforming to common sense.}
\end{itemize}
\paragraph{Listening}
\begin{itemize}
    \item \texttt{DIC:} the act of paying attention to hear.
    \item \texttt{ELO:} N.A.
    \item \texttt{CHAT:} (Question) How much did the user seem to pay attention to what you said?~\cite{see2019makes,sinha2020learning}
    \item \texttt{The quality of the response paying attention to the context.}
\end{itemize}
\paragraph{Maintain Context}
\begin{itemize}
    \item \texttt{DIC:} to continue the (context).
    \item \texttt{ELO:} N.A.
    \item \texttt{CHAT:} (Question) Does the response serve as a valid continuation of the preceding conversation?~\cite{mehri2020usr}
    \item \texttt{The quality of the response being a valid continuation of the context.}
\end{itemize}
\paragraph{Logic}
\begin{itemize}
    \item \texttt{DIC:} the study of reasoning thinking.
    \item \texttt{ELO:} the conformity of generative grammar.~\cite{lenci2004logic}
    \item \texttt{CHAT:} : the degree to which the post~(i.e., context) and the reply logically match.~\cite{li2018syntactically}
    \item \texttt{The quality of the response reasonably matching the context.}
\end{itemize}
\paragraph{Suggestions} It's worth mentioning that ``\texttt{Coherence}'' possesses two definitions. The definition of ``\texttt{Local Coherence}'' is the same as ``\texttt{Relevance}''. While the definition of ``\texttt{Global Coherence}'' is a criterion of the entire dialogue, making is  not suitable for the static human evaluation. Besides ``\texttt{Coherence}'', ``\texttt{Listening}'' also refers to ``\texttt{Relevance}''. Thus, we prefer to use the most commonly used ``\texttt{Relevance} as a standard criterion. Meanwhile, We also prefer to use ``\texttt{Consistency}'' as the second standard criterion in this group. As the ``\textit{context}'' belongs to ``\textit{the known information}, so that the definition of ``\texttt{Logic}'' can be covered by ``\texttt{Consistency}''. ``\texttt{Sensibleness} and ``\texttt{Maintain Context}'' are not selected because they refer to multiple criteria that we already select as standard ones. 
\subsection{Group 3}
\paragraph{Informativeness}
\begin{itemize}
    \item \texttt{DIC:} the quality of providing useful information.
    \item \texttt{ELO:} the amount of  new information contained in a text.
    \item \texttt{CHAT:} the quality of the response providing new information.
    \item \texttt{The quality of the response providing new information.}
\end{itemize}
\paragraph{Diversity}
\begin{itemize}
    \item \texttt{DIC:} the quality of being different.
    \item \texttt{ELO:} a rich mix of differences.
    \item \texttt{CHAT:} the quality of the response not repeating information of the context.
    \item \texttt{The quality of the response providing new information.}
\end{itemize}
\paragraph{Specificity}
\begin{itemize}
    \item \texttt{DIC:} the quality of being the only one thing.
    \item \texttt{ELO:} N.A.
    \item \texttt{CHAT:} the quality of the sensible response being specific to the context. For example, given a dialogue context ``\texttt{Person A: I love tennis!}'', the response ``\texttt{Person B: Me too. I can't get enough of Roger Federer!}'' is more specific than the response ``\texttt{That's nice.}''~\cite{danie2020towards}
    \item \texttt{The quality of the sensible response being specific to the context.}
\end{itemize}
\paragraph{Proactivity}
\begin{itemize}
    \item \texttt{DIC:} the quality of acting in advance of  a future situation rather than reacting.
    \item \texttt{ELO:} the quality of taking initial change.
    \item \texttt{CHAT:} the quality of the system to move the dialogue to new topics~(as shown in Figure~\ref{fig:proactivity}).
    \item \texttt{The quality of the system to move the dialogue to new topics.}
    \begin{figure}
        \centering
        \includegraphics[width=\textwidth]{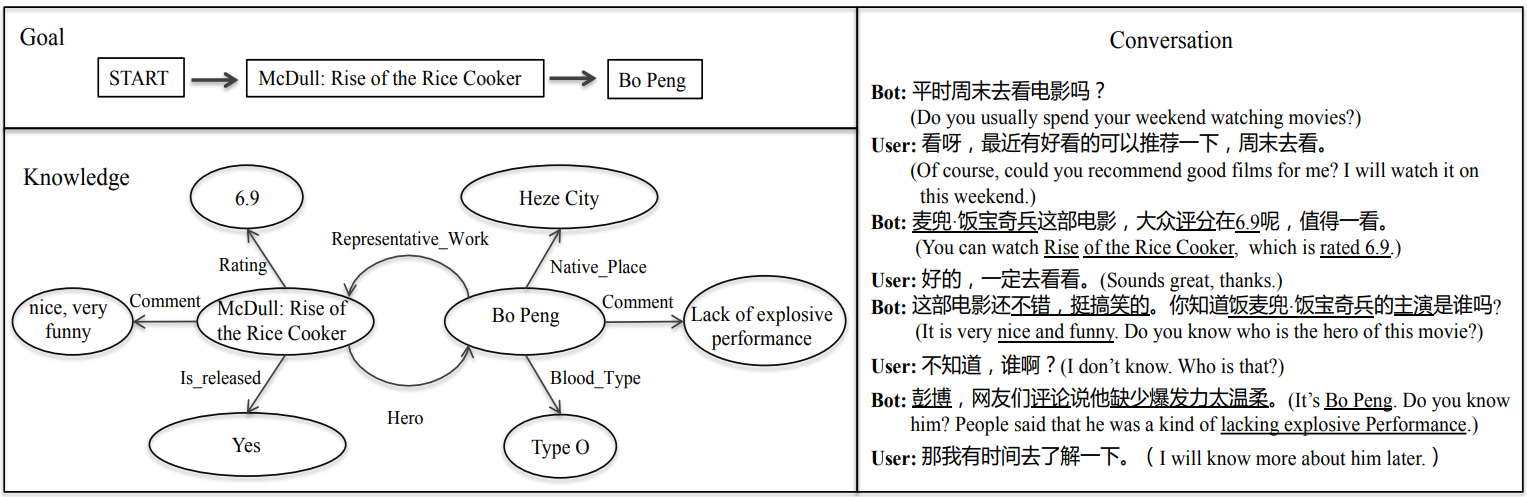}
        \caption{An example of Proactivity in~\cite{wu2019proactive}: the dialogue is moving from one topic~(McDull: Rise of the Rice Cooke) to another topic~(Bo Peng)}
        \label{fig:proactivity}
    \end{figure}
\end{itemize}
\paragraph{Flexible}
\begin{itemize}
    \item \texttt{DIC:} the capability of changing to suit new and variable situations.
    \item \texttt{ELO:} the ability of an individual to adapt to requirements.
    \item \texttt{CHAT:} (Question) Is the system flexible and adaptable to the user and their interests? \cite{mehri2020unsupervised}
    \item \texttt{The quality of the system adaptable to the user and their interests.}
\end{itemize}
\paragraph{Suggestions} We prefer to use ``\texttt{Informativeness}'' as a standard criterion in this group for three reasons. First, ``\texttt{Diversity}'' has the same definition with ``\texttt{Informativeness}'' but its frequency is less. Second, ``\texttt{Specificity}'' is the same with ``\texttt{Sensibleness}'', which is not selected because it can be represented by the standard criteria, i.e., ``\texttt{Readability}'', ``\texttt{Relevance}'', and ``\texttt{Consistency}''. Similarly, ``\texttt{Flexible}'' is not selected because ``\textit{the user and their interests}'' belongs to ``\textit{the known information}'', and thus it can be represented by ``\texttt{Consistency}''. Third, ``\texttt{Proactivity}'' is also not selected because it is designed for a special chatbot that requires a list of pre-defined topics~(e.g., START $\rightarrow$ cDull: Rise of the Rice Cooke $\rightarrow$ Bo Peng)~\cite{wu2019proactive}.
\subsection{Group 4}
\paragraph{Overall Quality}
\begin{itemize}
    \item \texttt{DIC:} N.A.
    \item \texttt{ELO:} N.A.
    \item \texttt{CHAT:} the overall impression of the response on multiple Criteria.
    \item \texttt{The overall impression of the response on multiple Criteria.}
\end{itemize}
\paragraph{Appropriateness}
\begin{itemize}
    \item \texttt{DIC:} the quality of being especially suitable.
    \item \texttt{ELO:} the quality of being particularly suitable.
    \item \texttt{CHAT:} the overall impression of the response on multiple Criteria.
    \item \texttt{The overall impression of the response on multiple Criteria.}
\end{itemize}
\paragraph{Humanness}
\begin{itemize}
    \item \texttt{DIC:} the quality of having human characteristics.
    \item \texttt{ELO:} the quality of being unique in contrast to other animals.
    \item \texttt{CHAT:} the likeliness of the response generated by a human.
    \item \texttt{The likeliness of the response generated by a human.}
\end{itemize}
\paragraph{Naturalness}
\begin{itemize}
    \item \texttt{DIC:} the quality of behaving in a normal way.
    \item \texttt{ELO:} the likeliness to be a nativelike selection of expression in a given context.
    \item \texttt{CHAT:} the plausibility of the response generated by a human.
    \item \texttt{The plausibility of the response generated by a human.}
\end{itemize}
\paragraph{Adequacy}
\begin{itemize}
    \item \texttt{DIC:} the quality of being enough in quality.
    \item \texttt{ELO:} N.A.
    \item \texttt{CHAT:} response is very reasonable.~\cite{liu2016not}
    \item \texttt{N.A.}
\end{itemize}
\paragraph{Suggestions} The definition of ``\texttt{Adequacy}'' in \texttt{DIC} is the same with ``\texttt{Sensibleness}'', which is not selected in \texttt{Group 2}. The definitions of ``\texttt{Adequacy}'' in \texttt{CHAT} and ``\texttt{Appropriateness}'' are the same with ``\texttt{Overall Quality}, which we find is rather vague for workers to judge. Several papers try to clear such criteria by defining them as ``natural, relevant and informative''~\cite{cui2019dal}, ``\textit{Coherence, language consistency, fluency and informativeness}''~\cite{cai2020data}, etc. This makes ``\texttt{Overall Quality}'' totally overlaps the other standard criteria. Thus, we prefer not to use ``\texttt{Overall Quality}'', `\texttt{Adequacy}'' or ``\texttt{Appropriateness}''. Both ``\texttt{Humanness}'' and ``\texttt{Naturalness}'' refer to ``\textit{the plausibility of a response generated by a human}''. However, ``\texttt{Humanness}'' also indicates to the human's uniqueness in contrast to other animals, which is more related to the chatbot than the response. As such, we prefer to use ``\texttt{Naturalness}'' as a standard criterion.
\subsection{Group 5}
\paragraph{Engagingness}
\begin{itemize}
    \item \texttt{DIC:} the quality to attract attention; the quality of being interesting.
    \item \texttt{ELO:} acknowledge readers' attention and connect to them.
    \item \texttt{CHAT:} the willingness to continue dialogue.
    \item \texttt{N.A.}
\end{itemize}
\paragraph{Interestingness}
\begin{itemize}
    \item \texttt{DIC:} the power of attracting people's attention.
    \item \texttt{ELO:} the quality of being interesting.
    \item \texttt{CHAT:} the willingness to continue dialogue.
    \item \texttt{N.A.}
\end{itemize}
\paragraph{Suggestions} ``\texttt{Engagingness}'' and ``\texttt{Interestingness}'' have the same definitions in both \texttt{DIC} and \texttt{CHAT}. However, we decide to select neither of them. As it is not possible for a worker to decide ``\texttt{Engagingness}'' or ``\texttt{Interestingness}'' from a response. Such criteria are more suitable for the self-play or interactive human evaluations that has multiple turns of interactions.
\subsection{Discussion}
In summary, we suggest to use the following five criteria in human evaluation for chatbots:
\begin{enumerate}[label={\bf Criteria \arabic*}]
    \item \texttt{Readability:} the quality of the response to be understood easily.
    \item \texttt{Relevance:} the quality of a response to connect with the context.
    \item \texttt{Consistency}: the quality of a response agreeing with the known information.
    \item \texttt{Informativeness:} the amount of new information in the response.
    \item \texttt{Naturalness:} the plausibility of the response generated by a human.
\end{enumerate}
We believe these standard criteria are enough to cover all finding criteria in off-the-shelf papers, and have non-overlapping in terms of definitions among themselves. Specifically, ``\texttt{Readability}'' describes the linguistic quality of the response, the evaluation of it has few connections to the dialogue context. ``\texttt{Relevance}'' describes the connection between the context and the response, especially, the topic connection. ``\texttt{Consistency}'' describes the conformity of the information in the response to the information in the context and what workers already know. ``\texttt{Informativeness}'' describes the amount of new details in the response with regard to the context. ``\texttt{Naturalness}'' is the only criteria that related to multiple aspects that needs the workers to make an overall judgement.

\bibliographystyle{unsrtnat}
\bibliography{ref}
\end{document}